\documentclass[conference]{IEEEtran}
\IEEEoverridecommandlockouts

\usepackage{cite}
\usepackage{amsmath,amssymb,amsfonts}
\usepackage{algorithmic}
\usepackage{microtype}
\usepackage{graphicx}
\usepackage{booktabs} 
\usepackage{enumitem}

\usepackage[table]{xcolor}
\usepackage{pgfplots}
\usepackage{pgfmath}
\usepackage{textcomp}
\usepackage{xcolor}
\usepackage{subcaption}
\usepackage{booktabs}
\PassOptionsToPackage{dvipsnames}{xcolor}
\usepackage{amsmath}
\usepackage{multirow}

\usepackage{tabularx}
\usepackage{makecell}
\usepackage{xfrac}
\usepackage{url}

\newcommand{\colorcorr}[1]{%
    \pgfmathparse{#1}%
    \pgfmathsetmacro{\corrvalue}{\pgfmathresult}%
    \edef\corrvalpt{\corrvalue pt} 
    \ifdim\corrvalpt>0.75pt \cellcolor{green!70} #1%
    \else\ifdim\corrvalpt>0.59pt \cellcolor{green!50} #1%
    \else\ifdim\corrvalpt>0.49pt \cellcolor{green!30} #1%
    \else\ifdim\corrvalpt<-0.50pt \cellcolor{red!40} #1%
    \else\ifdim\corrvalpt<-0.25pt \cellcolor{red!10} #1%
    \else #1%
    \fi\fi\fi\fi\fi
}

\def\BibTeX{{\rm B\kern-.05em{\sc i\kern-.025em b}\kern-.08em
    T\kern-.1667em\lower.7ex\hbox{E}\kern-.125emX}}

\begin{document}

\title{Will It Zero-Shot?:\\Predicting Zero-Shot Classification Performance For Arbitrary Queries}


\author{\IEEEauthorblockN{Kevin Robbins}
\IEEEauthorblockA{\textit{Computer Science} \\
\textit{George Washington University}\\
Washington, DC, USA \\
kevin.robbins@gwu.edu}
\and
\IEEEauthorblockN{Xiaotong Liu}
\IEEEauthorblockA{\textit{Computer Science} \\
\textit{George Washington University}\\
Washington, DC, USA \\
liuxiaotong2017@gwu.edu}
\and
\IEEEauthorblockN{Yu Wu}
\IEEEauthorblockA{\textit{Computer Science} \\
\textit{George Washington University}\\
Washington, DC, USA \\
yu.wu1@gwu.edu}
\and
\IEEEauthorblockN{Le Sun}
\IEEEauthorblockA{\textit{Computer Science} \\
\textit{George Washington University}\\
Washington, DC, USA \\
le.sun@email.gwu.edu}
\and
\IEEEauthorblockN{Grady McPeak}
\IEEEauthorblockA{\textit{Computer Science} \\
\textit{George Washington University}\\
Washington, DC, USA \\
gradymcpeak@gwu.edu}
\and
\IEEEauthorblockN{Abby Stylianou}
\IEEEauthorblockA{\textit{Computer Science} \\
\textit{Saint Louis University}\\
Saint Louis, USA \\
abby.stylianou@slu.edu}
\and
\IEEEauthorblockN{Robert Pless}
\IEEEauthorblockA{\textit{Computer Science} \\
\textit{George Washington University}\\
Washington, DC, USA \\
pless@gwu.edu}
}

\maketitle

\begin{abstract}
Vision-Language Models like CLIP create aligned embedding spaces for text and images, making it possible for anyone to build a visual classifier by simply naming the classes they want to distinguish. However, a model that works well in one domain may fail in another, and non-expert users have no straightforward way to assess whether their chosen VLM will work on their problem.  We build on prior work using text-only comparisons to evaluate how well a model works for a given natural‐language task, and explore approaches that also generate synthetic images relevant to that task to evaluate and refine the prediction of zero‐shot accuracy. We show that generated imagery to the baseline text‐only scores substantially improves the quality of these predictions.  Additionally, it gives a user feedback on the kinds of images that were used to make the assessment.  Experiments on standard CLIP benchmark datasets demonstrate that the image-based approach helps users predict, without any labeled examples, whether a VLM will be effective for their application.
\end{abstract}


\section{Introduction}
\label{introduction}

\begin{figure}[t]
    \centering
    \includegraphics[width=\columnwidth]{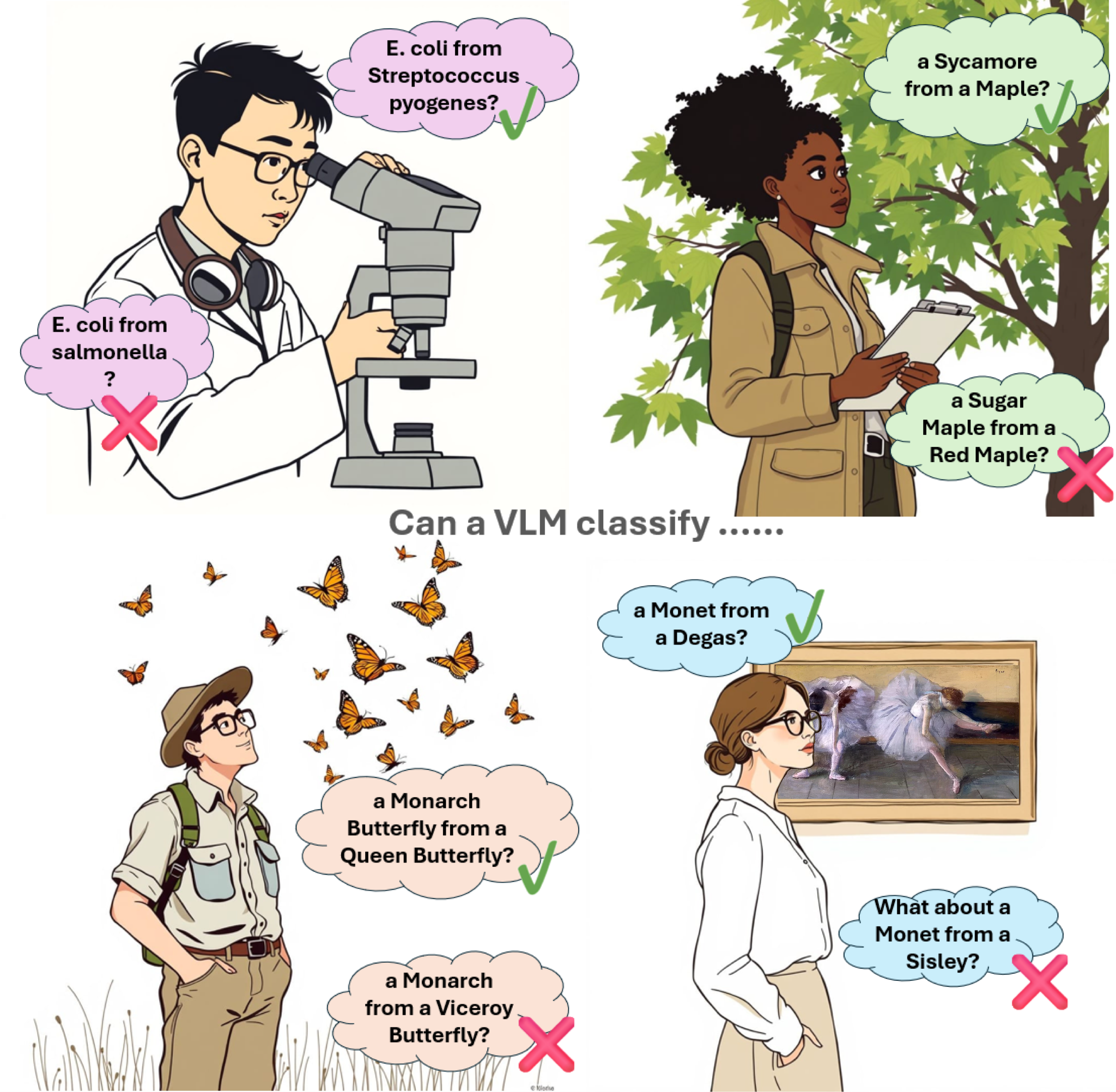}
    \caption{
Vision-Language Models (VLMs) enable users without machine learning expertise to rapidly deploy visual classifiers. These users need practical tools to determine if a VLM will be effective for their tasks. This paper introduces a simple approach for predicting zero-shot classification performance by analyzing the internal consistency of CLIP's text and image embedding spaces.
}
\label{fig:frontpage}
\end{figure}

Large-scale vision-language models (VLMs) that integrate text and imagery, such as CLIP~\cite{radford2021learning}, are accelerating the development of many types of visual algorithm in numerous application domains. These models are appealing because their behavior can be specified using natural language, making them accessible even to users without deep computer vision expertise. Creating classifiers simply by naming the classes that need to be distinguished, without any training or fine-tuning on new data, lowers the barrier to entry for visual classification. The initial CLIP paper showed that zero-shot performance can beat the state of the art across many domains. But just because a VLM based classifier can be easily defined with class names and works well in some cases, it does not guarantee the effectiveness for every classification problem. 

Approaches to evaluating VLM effectiveness for specific problems were recently proposed using a method comparing different VLM models to select the best model. The Language Only Visual Model selection (LOVM) work~\cite{zohar2023lovmlanguageonlyvisionmodel} predicts how well a collection of VLM models will perform on an arbitrary zero-shot task; their approach starts with a baseline prediction for each model based on measuring model accuracy on ImageNet, and then modifies that baseline score based on how well it classifies generated captions (used as a proxy for images).  By comparing many models this way, LOVM can pick the best VLM for a task given only a text description (e.g., distinguishing similar butterflies).  The primary motivation for LOVM is to select among a family of models to choose the best one for a problem.  We consider the scenario where users have a specific model and need to answer the question: "For the model I have access to, how well will it work on my task?". 
 Figure~\ref{fig:frontpage} shows motivating examples; in a wide variety of problem domains, subject matter experts may want to deploy zero-shot learning approaches and need to know if a model is likely to be effective.

Consider one possible task: "I have a picture of a butterfly, is it a Monarch or a Viceroy?".  VLMs may struggle with such a task for several reasons: (a) both the words "Monarch" and "Viceroy" have meanings unrelated to the butterfly, (b) Monarch and Viceroy butterflies are quite similar in appearance, (c) the data that the VLM is trained on may itself be mislabeled about these classes.  These are reasons why a user might want to ask if a given model would be effective for their task.  But making specific accuracy predictions may also be hard based on task descriptions: the user might have in mind butterflies pinned in a case where they can take well lit, high-resolution pictures, or they may have in mind smartphone images captured in scenarios with bad lighting and motion blur.


In this paper, we try to realistically address this scenario.  First, we improve the baseline approach to understanding how well a model understands concepts related to a given task using text captions as image proxies.  We demonstrate that evaluating VLM performance on task-relevant synthetic images improves that understanding compared to text-only analysis, with evaluation across many datasets from the CLIP benchmark suite. However comparisons to existing datasets are not useful to users, so we also built a user interface that can accept natural language queries and show users the generated images used in our calculations. Showing the user the synthetic images allows them to iterate over their queries to make them more specific, while also confirming that the images look like the data they intend to use in their application. This precision and model feedback are not necessarily possible with text as image proxies. Our specific contributions are:
\begin{itemize}
\item an improved algorithmic approach for predicting zero-shot VLM accuracy on a given class using only the class description;
\item a quantification of the performance advantage from using generated imagery in addition to text for prediction accuracy across CLIP benchmark datasets;
\item a web interface that deploys this zero-shot accuracy prediction approach, enabling users to compare models and refine queries without needing labeled data.
\end{itemize}


\section{Related Work}

\paragraph{Zero-Shot Learning}
Zero-shot learning (ZSL) seeks to classify images into classes that are described in some way but for which no training data is provided. The foundational work of~\cite{lampert2009learning} introduced and defined the ZSL problem. Early efforts to solve this problem involved the identification of attribute vectors and the composition of attributes by linear combinations of these vectors, as demonstrated in studies~\cite{romera2015embarrassingly,kodirov2017semantic,fu2015transductive}. Then these attribute vectors were applied in the inference phase to facilitate class prediction. With the development of deep learning, non-linear approaches for extracting attributes from images became possible, and ZSL approaches started to be more successful~\cite{wang2018zero}. Related zero-shot learning prediction methodologies to ours include LOVM~\cite{zohar2023lovmlanguageonlyvisionmodel} which attempted to create a benchmark for ranking VLMs for any given zero-shot task, and ModelGPT ~\cite{tang2024modelgptunleashingllmscapabilities} which is a hypernetwork intended to generate customized models faster than fine-tuning with just a user description.

 In predicting zero-shot and open-vocabulary performance of VLMs, Zero-Shot Out-of-Distribution (OOD) Detection \cite{Esmaeilpour2022} uses a set of generated seen and unseen labels to develop a score that predicts if a sample image is more similar to the unseen labels than the set of seen labels by doing cosine similarity for all labels and image embedding. These unseen labels are generated by a pre-trained image description generator that takes the CLIP image embedding as an input and outputs labels that are closer to the image in CLIP space.

\paragraph{CLIP}
Vision language models (VLM) such as the Contrastive Language Image Pre-Training (CLIP) model~\cite{radford2021learning} try to unify image and text space. Formally, CLIP tries to optimize an image encoder to map images into a set of features $I$, and a text encoder to map text into features $T$, maximizing the conditional log-probabilities for images and captions that are associated. A huge number of pairs of images and their associated captions are leveraged to train this model to have aligned image and text representations.

The CLIP framework can be used for classification, where the class name serves as the text prompt (e.g., ``an image of a 2003 Corvette''). In the standard approach, the predicted class for an image is determined by finding the class with the highest similarity between its text embedding and the embedding for the image. Building on this,~\cite{pratt2023does} improved the capabilities of VLMs for zero-shot learning by using more descriptive prompts rather than class names. By searching external vision-language databases with CLIP, CaSED~\cite{conti2023vocabulary} tries to solve the image zero-shot task without actual class names. Tip-Adapter~\cite{zhang2022tip} enhances CLIP's performance in few-shot settings while preserving the training-free advantage of zero-shot classification. There has also been work to improve zero-shot classification performance by improving the CLIP training data quality, including~\cite{fan2024improving} which used a large language model to rewrite the text descriptions for each CLIP training image.

Beyond these improvements in using CLIP for zero- and few-shot learning, there are a large number of extensions of CLIP (as well as other VLMs) that support visual question answering and captioning, such as SigLIP~\cite{zhai2023sigmoid}, FLAVA~\cite{singh2022flava}, and tools like DALL-E and its successors~\cite{dalle} that focus on image generation.

\paragraph{VLM Modality Gap}
In order to accurately predict whether CLIP represents a concept well or not, it is necessary to understand how the CLIP embedding space is organized. While the CLIP training objective tries to align related image and text representations, empirical observations from~\cite{zhou2023clip} and~\cite{liang2022mind} reveal a persistent modality gap within the shared latent space, where the image representations for a particular concept and the text representations for that concept are well clustered in their respective embedding spaces, but not well aligned across the embedding spaces. 

The existence of this modality gap suggests that approaches to evaluating whether CLIP understands a visual concept cannot be based simply on a threshold of how similar images and text descriptions of those images are. Instead, our proposed approach analyzes the internal consistency of the image and text model representations. The approach does not require labeled data, making it accessible to a variety of real-world end users, use cases, and problem domains.


\section{Method}
In this section, we describe our method to predict how well a CLIP-based model can classify an arbitrary class or concept defined in natural language. First, for a query class or concept, we create a list of related classes or concepts. This can be generated in several ways: provided by the end user (e.g., a medical team may already have a list of image types they care about); sourced from WordNet synsets; or generated by a large language model prompted with the query to generate related classes or concepts. For example, in a medical imaging context, we might be interested in how well a CLIP-variant classifies ``early stage tumors.'' The related concepts for this class might be ``benign growths'' or ``inflammation.'' The ideal model would be able to differentiate between these concepts, while models less suited to the task would not be.

Next, we consider the query text. Intuitively, if the VLM ``understands'' a concept, it should accurately distinguish examples of this query class from related but incorrect classes. While we do not have varying real-world examples of the query class -- it is only expressed in natural language -- we can simulate examples by using image and text generation tools to create diverse examples of the class. 

This idea draws inspiration from the cycle consistency losses used in unsupervised learning, which ensure that transformations applied to data can be reversed with minimal information loss. If we generate example images using a CLIP-based image generator, such as Stable Diffusion (which was trained using the OpenCLIP-ViT/H text-encoder)~\cite{rombach2022high}, map that image into the CLIP image embedding space, and then find similar features in the text embedding space, we expect that the textual description most closely related to the original query should be among the top matches. This consistency between the generated images and the text embeddings reflects the model’s ability to correctly capture the essence of the query. If the model is truly capable of understanding the specific concept, such as ``early-stage tumor,'' the generated images should map closely to relevant textual descriptions, and unrelated or incorrect classes (like ``benign growth'' or ``inflammation'') should be less similar in the embedding space. Using this cycle of generation and embedding, we can effectively gauge the accuracy of the model in interpreting and distinguishing the specified concept, even in the absence of labeled data.

We explore various image- and text-based scoring methods and analyze their correlation with true zero-shot accuracies on established labeled datasets. The following sections detail each step of this process.

This approach is primarily targeted for scenarios where labeled data is unavailable, difficult to obtain, or costly to generate. For well-labeled datasets, it is possible to simply measure the zero-shot classification accuracy. However, in many real-world applications, end users may need to determine if there exists a model that can support their task before investing in annotation efforts. Furthermore, the way a classification problem is framed in natural language can significantly impact the performance of a model. Subtle differences in wording -- such as describing a category as ``early-stage tumor'' versus ``precancerous growth'' -- can lead to variations in how well the model distinguishes between relevant concepts. By enabling users to explore different articulations of their task and assess how the model responds, our method provides a powerful tool for refining label definitions, even in already labeled datasets.

\subsection{Models}
In this paper, we evaluate our proposed methods on various Vision-Language Models (VLMs) inspired by the `CLIP' framework, which combines an image encoder and a text encoder to align visual and textual representations within a shared embedding space. Specifically, we consider:

\begin{itemize}
\item \textbf{CLIP}~\cite{radford2021learning}: The most popular initial VLM, integrating an image encoder and a text encoder. CLIP is trained to align image and text representations within a shared embedding space using a softmax-based contrastive learning loss.

\item \textbf{SigLIP}~\cite{zhai2023sigmoid}: An adaptation of the CLIP framework, but trained with a sigmoid-based loss instead of softmax, improving robustness to out-of-distribution data and long-tail categories.

\item \textbf{FLAVA}~\cite{singh2022flava}: A multi-modal framework that integrates an image encoder, a text encoder, and a multi-modality encoder designed for joint vision-language understanding. 

\end{itemize}

\subsection{Datasets}
While our proposed approach is designed for use cases with no labeled training data, we use labeled training data for evaluation purposes. In order to cover a broad range of image classes and settings, we selected 9 CLIP benchmark datasets, as well as CUB-200~\cite{CUB200} for more fine-grained classes. The CLIP-benchmark datasets that were selected are ImageNet~\cite{deng2009imagenet} and ObjectNet~\cite{barbu2019objectnet}, for their scale; CIFAR100~\cite{Krizhevsky2009learningmultiple}, for its focus on small images; Food101~\cite{bossard14}, FGVC-Aircraft~\cite{maji13fine-grained}, Flowers-102~\cite{Nilsback08}, Stanford Cars~\cite{krause2013cars}, and Oxford-IIIT Pets~\cite{parkhi2012cats} for their fine-grained classes;  and  RESISC-45~\cite{cheng2017remote}, for its focus on non-standard (satellite) imaging. 


\subsection{Generated Images}
For a given natural language query, we use SDXL-Lightning~\cite{lin2024sdxllightning} trained on LAION-5B~\cite{schuhmann2022laion} to generate twenty $1024 \times 1024$ pixel synthetic images of the class. The prompt for image generation is generated by GPT-4o to be a realistic caption of an image of a \{class\_name\} from the \{domain\}. For example, the domain for CUB-200 is `birds'. We first generated images with the simple prompt of "a photo of a \{class\_name\}, a type of \{domain\}", but found that the generated captions yielded more realistic and diverse pictures, whereas the same image prompt yielded very similar images. For a new query from an unknown domain, the user could specify the domain or it could be generated by an LLM (an approach we use on our web interface discussed below). Each image's corresponding text label is "a photo of a \{class\_name\}".



\subsection{Scores}
After generating the text descriptions the list of related classes, and the text and/or images for the query, we use CLIP-ViT-B/32 (https://huggingface.co/openai/clip-vit-base-patch32) to compute feature embeddings, and consider different scores that attempt to capture relevant information for predicting the zero-shot accuracy of the CLIP model. We describe each of the different scores in depth below.

\paragraph{Consistency Score}
The first score we propose uses the generated images and compares their embeddings to the text embeddings for the list of related classes. The point of this score is to measure the consistency of embedding shifts for image and text embeddings of one class to a similar neighbor class. To get a single score for class $i$, we minimize over all other classes, to focus on the most inconsistent (confusing) alternative class. Specifically, we define a consistency score, $s_k$:
\begin{equation}
 s_k = \min_{j \neq i} \cos{(I_{i_k}-\overline{I}_j,T_i-T_j)} 
    \label{equ:score}
\end{equation}
where $I_{i_k}$ represents the image embedding of the $k$-th generated image belonging to class $i$, while $\overline{I}_j$ denotes the mean vector of all generated image embeddings for class $j$. We define $S_{i_{cs}}$ to be the consistency score for class $i$ computed as the average over all the images in the class. Each difference computed represents the change in text or image embeddings from class $i$ to class $k$.

\paragraph{Multimodal Silhouette Score.}
The second score we propose focuses on how classification is often a task of distinguishing a class from its nearest neighbor. This score takes inspiration from the silhouette score \cite{SilhouettesRousseeuw} that was used for the granularity scores by LOVM~\cite{zohar2023lovmlanguageonlyvisionmodel} while also updating it to account for the multimodal nature of vision-language models. The generated images are treated as a cluster in a class's image space. The multimodal silhouette score evaluates that cluster, its modality gap, and its distance from its nearest neighbor (weighted by $\lambda=2.5$). For each class \(i\), let \(\{I_{\!i_k}\}_{k=1}^{N_i}\) be its images, and \(T_i\) its text embedding.
We define the intra-class distances for each image \(I_{\!i_k}\) as:

\begin{equation}
a\bigl(I_{i_k}\bigr)
= 1 - \cos\!\bigl(I_{i_k}, \overline{I}_{i}\bigr)
   + \lambda\,\Bigl[\,1 - \cos\!\bigl(I_{i_k}, T_{i}\bigr)\Bigr],
   \label{eq:a}\\[6pt]
\end{equation}   
and the inter-class distances as:
\begin{align}   
b\bigl(I_{i_k}\bigr)
&= \min_{j \neq i}\!
   \biggl\{
       \Bigl[\,1 - \cos\!\bigl(I_{i_k}, \overline{I}_j\bigr)\Bigr] 
     \notag\\
&\qquad\quad
     + \lambda\,\Bigl[\,1 - \cos\!\bigl(I_{i_k}, T_j\bigr)\Bigr]
   \biggr\}.
   \label{eq:b}
\end{align}

Here, \(\lambda\) balances how much we weigh image-text proximity versus image-image cohesion.
The multimodal silhouette for \(I_{\!i_k}\) is then defined in a silhouette-like manner:
\[
  \mathrm{sil}_{\mathrm{mm}}\!\bigl(I_{\!i_k}\bigr)
  \;=\;
  \frac{\,b(I_{\!i_k}) - a(I_{\!i_k})\,}
       {\max\!\bigl\{\,a(I_{\!i_k}),\,b(I_{\!i_k})\bigr\}}
\]
A larger value means \(I_{\!i_k}\) is well separated from other classes in both image and text spaces. We define $S_{i_{sil}}$ as the multimodal silhouette score for class $i$ computed as the average for all images in the class. This score acts as a classification margin that is more sensitive to the intra-class distances of a group of images. The final multimodal silhouette score and consistency scores are presented as coefficients to evaluate the performance of the VLM in this specific class $i$. 

\paragraph{Compound Score}
Finally, we propose a compound approach that considers both the internal consistency of image and text embeddings as well as the intra- and inter-class distances. The compound score
uses a scaling factor ($\alpha$ = 4) chosen to approximately scale the effect of the silhouette score to be the same as the consistency score. This score measures the consistency for every alternative class, and penalizes a small or negative silhouette score:
\[
  S_i
  \;=\;
    S_{i_{cs}} + \alpha {S_{i_{sil}}}.
\]

\begin{figure*}[t!]
    \centering
    \begin{subfigure}[b]{0.245\textwidth}
        \includegraphics[width=1\textwidth]{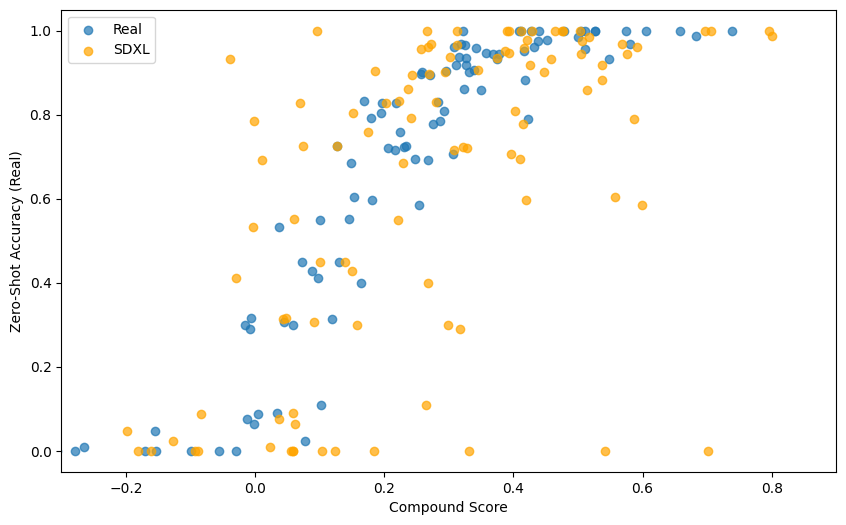}
        \caption{Flowers-102}
    \end{subfigure}
    \begin{subfigure}[b]{0.245\textwidth}
        \includegraphics[width=1\textwidth]{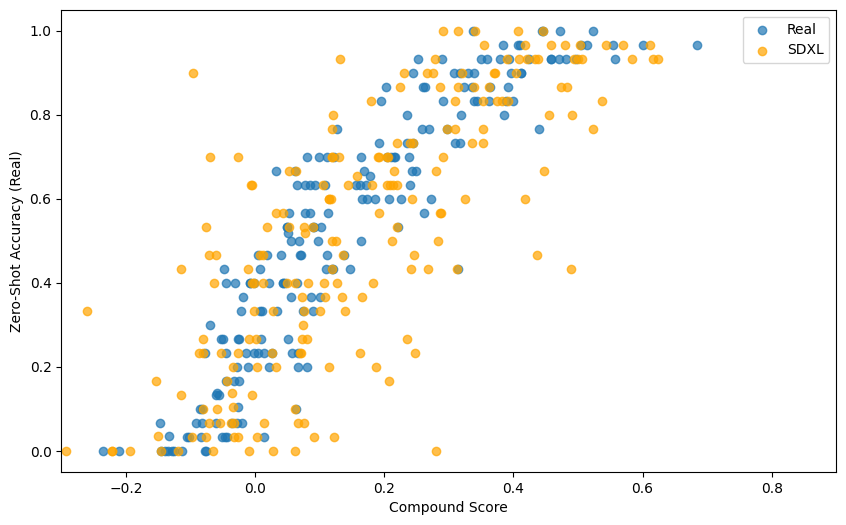}
        \caption{CUB-200}
    \end{subfigure}
    \begin{subfigure}[b]{0.245\textwidth}
        \includegraphics[width=1\textwidth]{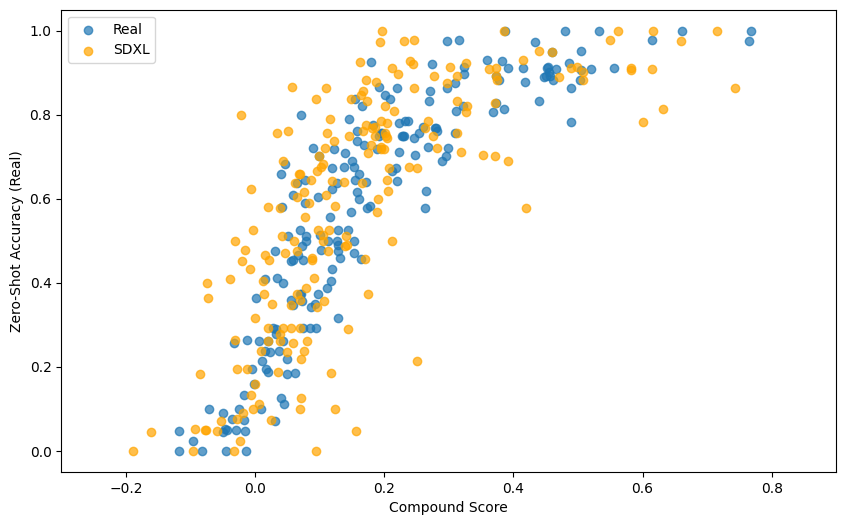}
        \caption{Stanford Cars}
    \end{subfigure}
    \begin{subfigure}[b]{0.245\textwidth}
        \includegraphics[width=1\textwidth]{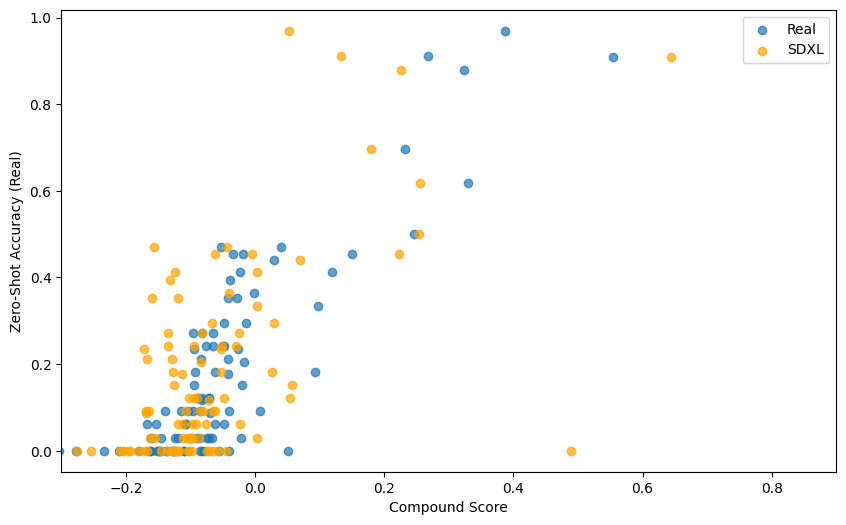}
        \caption{FGVC-Aircraft}
    \end{subfigure}
    \caption[VLM-Evaluation: Correlation between consistency score with accuracy]{Each scatter plot shows the relationship between zero-shot classification accuracy on the real dataset and our compound score in real (blue) or generated (orange) data. Each dot represents a class in datasets.}
    \label{fig:scatter}
\end{figure*}

\paragraph{Text-Based Scores}
We also consider a purely text-based score that does not require the generation of any synthetic images. This method leverages detailed textual descriptions as proxies for images to eliminate the need for image generation. To formulate our text-only consistency score, we define two types of prompts:

\begin{enumerate}
    \item \textbf{Plain Text:} "a photo of a \{*class\_name*\}”
    \item \textbf{Descriptive Text:} "a photo of a \{*likely descriptive caption for class\_name including the exact class\_name*\}”
\end{enumerate}

Here, the descriptive text is generated using GPT-4o. For instance, for the class ``Golden Retriever,'' examples of the descriptive text are ``a photo of a large golden retriever with short, silky fur'' and ``a photo of a medium golden retriever with long, wavy fur.'' Using these descriptions allows us to generate diverse images for a given class. More examples can be found in the Appendix Section~\ref{sec:imageCaptioning}.


\paragraph{Text-Only Consistency Score.}
To mirror the image-based consistency score, we replace synthetic images with descriptive text embeddings. For each class~$i$, let $T_{i_k}^d$ be the CLIP embedding of the $k$-th descriptive text, and let $T_i$ be the embedding of the plain text prompt (e.g., ``a photo of a \{\texttt{class\_name}\}''). We compute:
\begin{equation}
\centering
\begin{array}{l}
 \displaystyle
 s_k^T = \min_{j \neq i} \cos\bigl(T_{i_k}^{d}-\overline{T}_j^{d},\,T_i - T_j\bigr),
 \label{equ:text-cs}
\end{array}
\end{equation}
where $\overline{T}_j^{d}$ is the mean of the descriptive text embeddings for class~$j$, and $T_j$ is the plain text embedding for class~$j$. We define $S_{i_{\mathrm{cs}}}^{\,T}$ to be the text consistency score for class $i$ computed as the average over all the images in the class.

\paragraph{Text-Based Silhouette Score.}
We also adapt the multimodal silhouette score for text-only data. For class~$i$, let $\{T_{i_k}^d\}_{k=1}^{N_i}$ be the descriptive text embeddings, and let $T_i$ be the plain text embedding. We define the intra-class term
\begin{equation}
  a(T_{i_k}^d)
  \;=\;
    1 - \cos\bigl(T_{i_k}^d,\overline{T}_i^d\bigr)
  \;+\;
  \lambda\,\Bigl[1 - \cos\bigl(T_{i_k}^d,T_i\bigr)\Bigr],
\end{equation}
and the nearest inter-class term:
\begin{eqnarray}
b\bigl(T_{i_k}^d\bigr)
&=&
\min_{j \neq i}
\Bigl\{
  \bigl[1 - \cos\bigl(T_{i_k}^d,\overline{T}_j^d\bigr)\bigr]
  \;+\;\nonumber\\
&&
  \lambda \,\Bigl[1 - \cos\bigl(T_{i_k}^d, T_j\bigr)\Bigr]
\Bigr\}.
\end{eqnarray}

where $\overline{T}_i^d$ is the mean of the descriptive text embeddings for class~$i$, and $\lambda=2.5$ balances plain-descriptive distances. Then the text-based silhouette is
\begin{equation}
  \mathrm{sil}_{\mathrm{text}}(T_{i_k}^d)
  =
  \frac{b(T_{i_k}^d) - a(T_{i_k}^d)}{\max\{a(T_{i_k}^d),\,b(T_{i_k}^d)\}},
\end{equation}
We define $S_{i_{\mathrm{sil}}}^{\,T}$ to be the text silhouette score for class $i$ computed as the average over all the images in the class.

\paragraph{Text-Based Compound Score.}
Finally, we combine the text-only consistency score and the text-based silhouette score:
\begin{equation}
  S_i^T
  \;=\;
    S_{i_{\mathrm{cs}}}^{\,T} \;+\; \alpha\,S_{i_{\mathrm{sil}}}^{\,T},
\end{equation}
where we set $\alpha=4$, so that its scale is matched to the image-based compound score.

\section{Experimental Results}
In this section, we demonstrate the efficacy of our proposed method and report results using the ten labeled datasets discussed in the methods section for evaluation purposes.  

\subsection{Comparison of Scores and Zero-shot Accuracy}
\setlength{\tabcolsep}{4pt} 
\begin{table*}[t]
    \centering
    \begin{minipage}{0.4\textwidth}
        \centering
        \caption{Zero-Shot Classification\\Accuracy for Real Images}
        \label{tab:real_image_accuracy}
        \resizebox{\textwidth}{!}{%
        \begin{tabular}{|l|ccc|}
        \hline
        \textbf{Dataset} & \textbf{CLIP} & \textbf{SigLIP} & \textbf{FLAVA} \\ \hline
        CUB      & \colorcorr{.52}  & \colorcorr{.73}  & \colorcorr{.45}  \\
        Cars     & \colorcorr{.58}  & \colorcorr{.93}  & \colorcorr{.29}  \\
        Aircraft & \colorcorr{.17}  & \colorcorr{.41}  & \colorcorr{.12}  \\
        Food     & \colorcorr{.82}  & \colorcorr{.94}  & \colorcorr{.79}  \\
        ImageNet & \colorcorr{.62}  & \colorcorr{.79}  & \colorcorr{.55}  \\
        ObjectNet & \colorcorr{.43} & \colorcorr{.65}  & \colorcorr{.33}  \\
        CIFAR    & \colorcorr{.64}  & \colorcorr{.71}  & \colorcorr{.64}  \\
        Flowers  & \colorcorr{.65}  & \colorcorr{.71}  & \colorcorr{.64}  \\
        Pets     & \colorcorr{.80}  & \colorcorr{.92}  & \colorcorr{.63}  \\
        RESISC   & \colorcorr{.51}  & \colorcorr{.61}  & \colorcorr{.44}  \\ \hline
        \end{tabular}
        
        }
    \end{minipage}%
    \hfill
    \begin{minipage}{0.59\textwidth}
    \centering
    \caption{Spearman Correlations for Compound Score \& Real Zero Shot Accuracy for Text and Image Methods (CLIP+SDXL-Lightning)}\label{tab:image_results}
    \vspace{-.01in}
    \resizebox{\textwidth}{!}{%
    \begin{tabular}{|l|cc|cc|}
        \hline
        \textbf{Dataset} & \multicolumn{2}{c|}{\textbf{Text Score}} & \multicolumn{2}{c|}{\textbf{Image Score}} \\ \cline{2-5}
                         & Compound      & Zero Shot       & Compound       & Zero Shot      \\ \hline
        CUB            & \colorcorr{.52} & \colorcorr{.33} & \colorcorr{.77} & \colorcorr{.57} \\
        Cars           & \colorcorr{.52} & \colorcorr{.27} & \colorcorr{.79} & \colorcorr{.74} \\
        Aircraft       & \colorcorr{.36} & \colorcorr{.32} & \colorcorr{.54} & \colorcorr{.55} \\
        Food           & \colorcorr{.34} & \colorcorr{.15} & \colorcorr{.25} & \colorcorr{.28} \\
        ImageNet       & \colorcorr{.29} & \colorcorr{.09} & \colorcorr{.50} & \colorcorr{.43} \\
        ObjectNet      & \colorcorr{.01} & \colorcorr{.01} & \colorcorr{.25} & \colorcorr{.25} \\
        CIFAR          & \colorcorr{-.10} & \colorcorr{-.04} & \colorcorr{.60} & \colorcorr{.56} \\
        Flowers        & \colorcorr{.60} & \colorcorr{.31} & \colorcorr{.60} & \colorcorr{.65} \\
        Pets           & \colorcorr{.37} & \colorcorr{.39}  & \colorcorr{.44} & \colorcorr{.49} \\
        RESISC         & \colorcorr{.34} & \colorcorr{.25} & \colorcorr{.55} & \colorcorr{.59} \\ \hline
    \end{tabular}
    }
\end{minipage}

\end{table*}
\begin{table*}[ht]
    \centering
    \begin{minipage}{0.3\textwidth}
        \centering
        \caption{Spearman Correlation for Image-based Scores\\(Model $\times$ Score $\times$ Image Generator)}\label{tab:combined_dalle_results_by_dataset}
        \resizebox{.99\textwidth}{!}{%
        \begin{tabular}{l|ccc|}
        \cline{2-4}
        \multicolumn{1}{l|}{}  & \multicolumn{3}{c|}{Generated with DALL-E 3} \\ \cline{1-4}
        \multicolumn{1}{|l|}{Dataset}  & CLIP & SigLIP & FLAVA \\ \hline
        \multicolumn{1}{|l|}{CUB}      & \colorcorr{.68}  & \colorcorr{.72}    & \multicolumn{1}{c|}{\colorcorr{.63}}    \\
        \multicolumn{1}{|l|}{Aircraft} & \colorcorr{.55}  & \colorcorr{.79}    &  \multicolumn{1}{c|}{\colorcorr{.59}}    \\
        \multicolumn{1}{|l|}{CIFAR}    & \colorcorr{.70}  & \colorcorr{.63}    & \multicolumn{1}{c|}{\colorcorr{.68}}     \\
        \multicolumn{1}{|l|}{RESISC}   & \colorcorr{.59}  & \colorcorr{.35}    &  \multicolumn{1}{c|}{\colorcorr{.65}}    \\ \hline
        \multicolumn{1}{l|}{}           & \multicolumn{3}{c|}{Compound Score} \\  \hline      
        \multicolumn{1}{|l|}{CUB}       & \colorcorr{.77}  & \colorcorr{.74}  & \colorcorr{.73}   \\
        \multicolumn{1}{|l|}{Aircraft}  & \colorcorr{.54}  & \colorcorr{.66}  & \colorcorr{.60} \\
        \multicolumn{1}{|l|}{CIFAR}     & \colorcorr{.60}  & \colorcorr{.53}  & \colorcorr{.59} \\
        \multicolumn{1}{|l|}{RESISC}    & \colorcorr{.55}  & \colorcorr{.25}  & \colorcorr{.51}   \\ \hline
        \cline{2-4}
        \multicolumn{1}{l|}{}  & \multicolumn{3}{c|}{Generated with SDXL} \\ \cline{2-4} 
        \end{tabular}
        }
    \end{minipage}
    \hfill
    \begin{minipage}{0.69\textwidth}
        \centering
        \caption{Spearman Correlation for Image-based Scores}
        \label{tab:image_results}
        \resizebox{\textwidth}{!}{%
        \begin{tabular}{l|ccccccccc|}
        \cline{2-10}
        & \multicolumn{3}{c|}{Consistency} & \multicolumn{3}{c|}{Silhouette} & \multicolumn{3}{c|}{Compound} \\ \hline
        \multicolumn{1}{|l|}{Dataset} & CLIP & SigLIP & \multicolumn{1}{c|}{FLAVA} & CLIP & SigLIP & \multicolumn{1}{c|}{FLAVA} & CLIP & SigLIP & FLAVA \\ \hline
        \multicolumn{1}{|l|}{CUB}       & \colorcorr{0.73}  & \colorcorr{0.73}  & \multicolumn{1}{c|}{\colorcorr{0.62}}   & \colorcorr{0.75}  & \colorcorr{0.68}  & \multicolumn{1}{c|}{\colorcorr{0.70}}   & \colorcorr{0.77}  & \colorcorr{0.74}  & \colorcorr{0.73}  \\
        \multicolumn{1}{|l|}{Cars}      & \colorcorr{0.69}  & \colorcorr{0.54}  & \multicolumn{1}{c|}{\colorcorr{0.65}}   & \colorcorr{0.79}  & \colorcorr{0.59}  & \multicolumn{1}{c|}{\colorcorr{0.81}}   & \colorcorr{0.79}  & \colorcorr{0.60}  & \colorcorr{0.80}  \\
        \multicolumn{1}{|l|}{Aircraft}  & \colorcorr{0.57}  & \colorcorr{0.56}  & \multicolumn{1}{c|}{\colorcorr{0.40}}   & \colorcorr{0.49}  & \colorcorr{0.62}  & \multicolumn{1}{c|}{\colorcorr{0.62}}   & \colorcorr{0.54}  & \colorcorr{0.66}  & \colorcorr{0.60}  \\
        \multicolumn{1}{|l|}{Food}      & \colorcorr{0.19}  & \colorcorr{0.27}  & \multicolumn{1}{c|}{\colorcorr{0.15}}   & \colorcorr{0.26}  & \colorcorr{0.20}  & \multicolumn{1}{c|}{\colorcorr{0.39}}  & \colorcorr{0.25}  & \colorcorr{0.22}  & \colorcorr{0.36} \\
        \multicolumn{1}{|l|}{ImageNet}  & \colorcorr{0.47}  & \colorcorr{0.47}  & \multicolumn{1}{c|}{\colorcorr{0.62}}   & \colorcorr{0.49}  & \colorcorr{0.46}  & \multicolumn{1}{c|}{\colorcorr{0.66}}   & \colorcorr{0.50}  & \colorcorr{0.48}  & \colorcorr{0.68}  \\
        \multicolumn{1}{|l|}{ObjectNet} & \colorcorr{0.19} & \colorcorr{0.08} & \multicolumn{1}{c|}{\colorcorr{0.21}}   & \colorcorr{0.25}  & \colorcorr{0.21} & \multicolumn{1}{c|}{\colorcorr{0.35}}  & \colorcorr{0.25}  & \colorcorr{0.16} & \colorcorr{0.34} \\
        \multicolumn{1}{|l|}{CIFAR}     & \colorcorr{0.55} & \colorcorr{0.43} & \multicolumn{1}{c|}{\colorcorr{0.53}}  & \colorcorr{0.60} & \colorcorr{0.57} & \multicolumn{1}{c|}{\colorcorr{0.59}}   & \colorcorr{0.60} & \colorcorr{0.53} & \colorcorr{0.59}  \\
        \multicolumn{1}{|l|}{Flowers}   & \colorcorr{0.63}  & \colorcorr{0.30}  & \multicolumn{1}{c|}{\colorcorr{0.44}}   & \colorcorr{0.58}  & \colorcorr{0.27}  & \multicolumn{1}{c|}{\colorcorr{0.64}}   & \colorcorr{0.60}  & \colorcorr{0.27}  & \colorcorr{0.62}  \\
        \multicolumn{1}{|l|}{Pets}      & \colorcorr{0.45}  & \colorcorr{0.48}  & \multicolumn{1}{c|}{\colorcorr{0.49}}   & \colorcorr{0.42}  & \colorcorr{0.50} & \multicolumn{1}{c|}{\colorcorr{0.41}}  & \colorcorr{0.44}  & \colorcorr{0.56}  & \colorcorr{0.49} \\
        \multicolumn{1}{|l|}{RESISC}    & \colorcorr{0.40}  & \colorcorr{0.27}  & \multicolumn{1}{c|}{\colorcorr{0.27}}   & \colorcorr{0.53}  & \colorcorr{0.23}  & \multicolumn{1}{c|}{\colorcorr{0.54}}   & \colorcorr{0.55}  & \colorcorr{0.25}  & \colorcorr{0.51}  \\ \hline
        \end{tabular}
        }
    \end{minipage}
\end{table*}


For each class in the datasets, we compute our image and text-based  scores. We evaluate using labeled datasets, so we can report on the zero-shot accuracy metric for each of these datasets.  

\paragraph{Image-Based Scores} Figure~\ref{fig:scatter} shows the relationship between our compound score and the true accuracy for each class in four of the datasets. The y-axis represents the zero-shot classification accuracy on the real dataset, and the x-axis represents the compound score for each class. 
To gain a qualitative understanding of the reasonableness of the generated images, each class in each dataset is represented with two points. Blue markers denote the scores computed based on embeddings of the real images in the test datasets, while orange markers denote generated images (that would be available in arbitrary circumstances). The figure shows a positive correlation between the classification accuracy and the scores computed with both the real image embeddings and the embeddings of the generated images, showing that the generated images serve as a good (but not perfect) proxy for the real images.

To quantify the relationship, we compute the Spearman correlation coefficient between the scores for the generated images and the zero-shot accuracy performance on real datasets to assess how strongly our scores are related to the true zero-shot classification accuracy. 
 Table~\ref{tab:real_image_accuracy}, shows the zero-shot classification accuracy for the real images in each of the datasets, when using the CLIP, SigLIP and FLAVA models.  These are the actual accuracies for Zero-Shot classification with these VLMs on images from these standard datasets.  The next three tables capture our prediction performance, how well our scores correlate with the per-class classification accuracy.

\paragraph{Compound Score vs. Zero-Shot}
In Table~\ref{tab:image_results}, we compare our compound score with a baseline approach of simply measuring the zero-shot classification performance on the generated images.  We observe that the compound score has higher correlation with the actual zero-shot learning accuracy of the real world data on a significant majority of the datasets, and also that computing these scores based on image embeddings is much better than computing these scores based on text embeddings.  We believe the improvements from the compound score arise because the consistency score and silhouette score capture useful measurements of the image clustering and local organization of similar classes in the embeddings spaces, which generalizes better than the classification accuracy of the generated images.

For some classes, the real image classification variations between classes have low correlations with both the text and image-based scores.  For some datasets like Food and PETS, the zero-shot classification is very accurate for most classes, making the rank-based correlation that Spearman correlation not a stable measure.  The other classes with very poor performance are in ObjectNet, which explicitly has images of objects that are more obscured in the scene and in unnatural positions (like a "rocking chair" image showing a small rocking chair laying sideways on a bed, taking up a small part of the image).  In this case, correlations are low because synthetic images (which are reasonable images of the class) are not representative of the images in the task.

\paragraph{Choice of image generator}
Table~\ref{tab:combined_dalle_results_by_dataset} shows the correlation of our
image-based scores for images generated using the DALL-E 3 image generator~\cite{dalle} (top), and the open-source SDXL-Lightning image generator~\cite{lin2024sdxllightning} (bottom). We run this experiment on a subset of the datasets due to the cost (in both dollars and time) of the DALL-E 3 image generator. In all of our result tables, entries that have correlations greater than 0.5 are color-coded in green, with higher correlations having brighter coloring.  The figure shows that both image generation models give scores with reasonably high correlations.  While Dall-E 3 images are often more compelling and realistic, the results for our purposes are similar.  Unless otherwise noted, the remainder of the paper uses the SDXL-Lightning model.



\paragraph*{Text-Based Scores.} Table~\ref{tab:text_results} shows the correlation for each of the different text-based scores, using each of the different models, across each of the datasets. These correlations are significantly lower than using the image-based approaches seen in Table~\ref{tab:image_results}, and in some cases are even negative. This suggests that simply using text prompts to ascertain how well a model can perform zero-shot learning for image datasets is insufficient.

\paragraph{Outlier Classes}
While the tables above present aggregate information computed over all of the classes in a dataset, we can also investigate the performance for specific classes. In particular, there are some exceptional outlier classes, where our score and the actual accuracy diverge.  These are points that are in the bottom-right or top-left corners of Figure~\ref{fig:scatter}.

\begin{figure}[t]
    \centering    
    \begin{subfigure}[b]{\columnwidth}
        \includegraphics[width=\columnwidth]{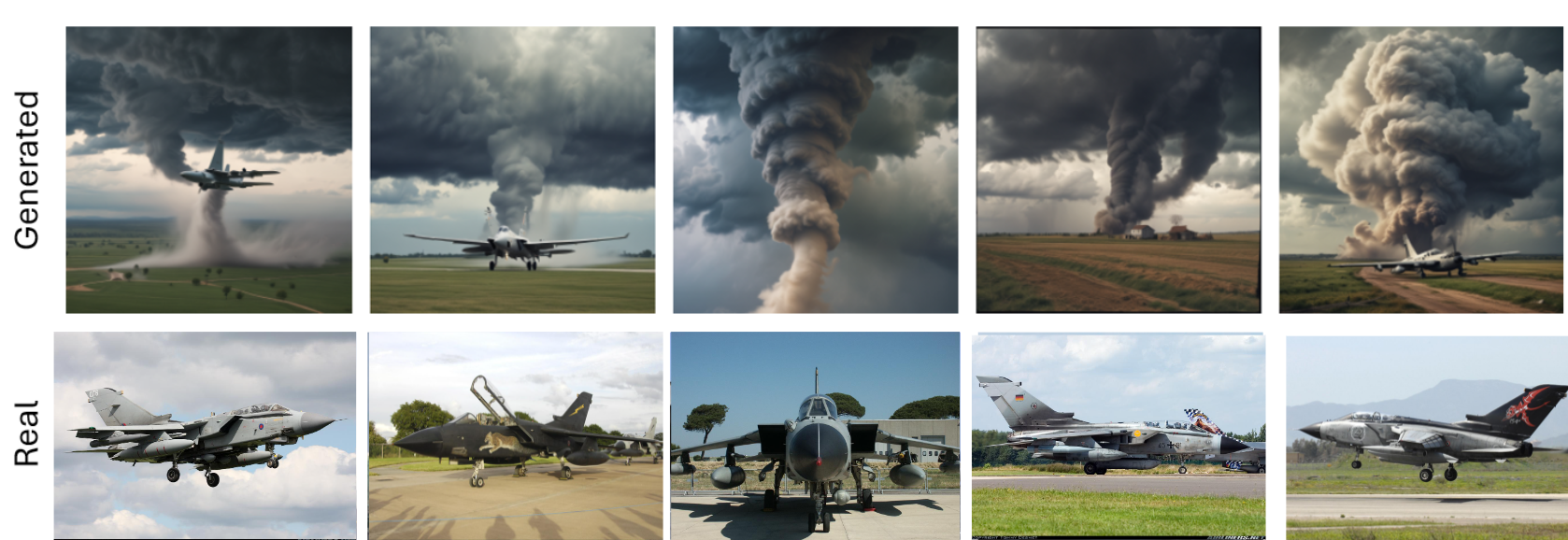}
        \caption{Aircraft:"Tornado"}
        \label{fig:bad_tornado}
    \end{subfigure}
    \begin{subfigure}[b]{\columnwidth}
        \includegraphics[width=\columnwidth]{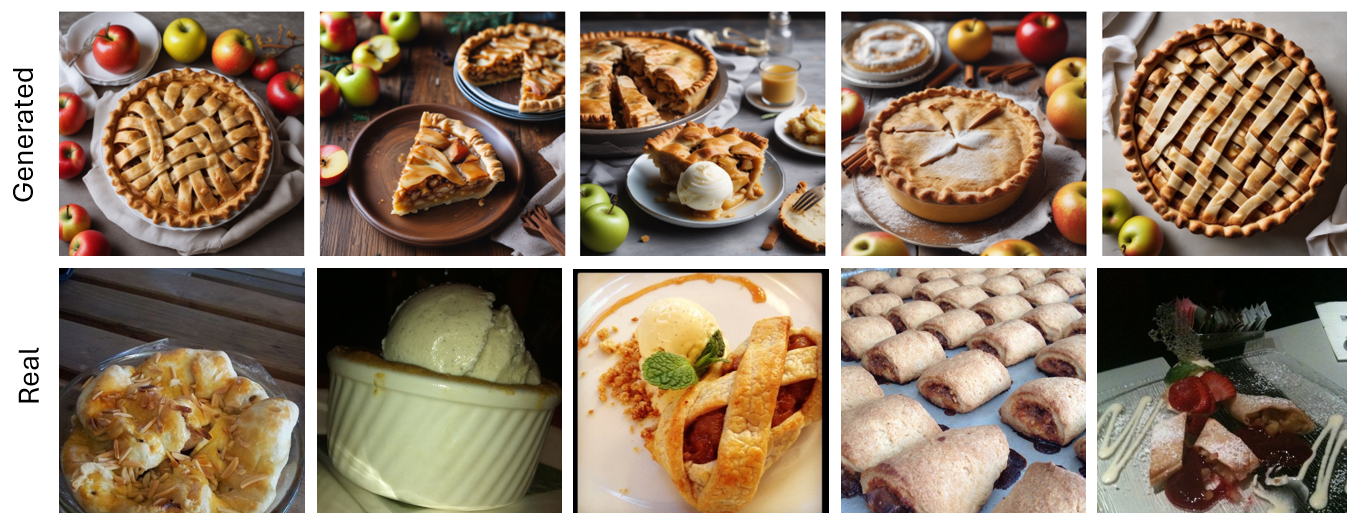}
        \caption{Food-101:"Apple Pie"}
        \label{fig:bad_apple}
    \end{subfigure}
    \caption{Example images from classes where the compound scores do not match the CLIP Zero-Shot Accuracy. These outliers can be caused by real dataset images being weaker depictions of the class than the generated images (a) or image generation misunderstanding the class (b).}
    \label{fig:bad_generation}
\end{figure}

One such outlier was the ``Tornado" class from the Aircraft dataset, shown in Figure~\ref{fig:bad_tornado}, where the top row shows generated images and the bottom row shows real images. Although the real images had very high zero-shot classification accuracy, our prediction on the generated images was quite low. The reason for this is that the image generation process confused the meaning of Tornado the aircraft, with the natural disaster. This suggests a potential limitation of our method: it may be challenging to generate images for classes that have somewhat ambiguous meanings, but that is irrelevant to the classification performance for real world images of the class. This can, however, be mitigated by a user visually confirming that the generated images are acceptable and regenerating them if not, potentially using modified image generation prompts (e.g., `a tornado, a type of aircraft').

Another interesting outlier class was ``Apple Pie" from the Food-101 dataset, shown in Figure~\ref{fig:bad_apple}. For this dataset, the generated apple pie pictures were all very clear depictions of the class and, since there are no other pies among the food classes, the apple pie class scores very well. The real apple pie pictures, on the other hand, score very poorly. When reviewing the images it is clear that the actual dataset images are more often deconstructed pastry plates, smaller handheld apple pastries,  or just more focused on apple pie dish accessories than the actual pie itself. This is not an issue of the generated images not being realistic depictions of apple pie, but more of the real images having a much looser interpretation of what counts as apple pie. In this way our method can be used to evaluate the quality of class depictions comparatively to the generated images. This might be useful to someone wanting to evaluate the labels of a dataset or understand why some set of data does not work well for their retrieval queries.

\section{Interactive Tool}
We finally present a web accessible interactive tool, ``Will CLIP Zero-Shot?''. This tool demonstrates our methods by allowing users to input any natural language query and see our prediction of the zero-shot classification accuracy for that query with CLIP. We describe a simple approach to convert our scores to a prediction of classification accuracy in Appendix Section~\ref{sec:scoreConversion}. This interactive tool can be accessed at \url{https://willitzeroshot.com}.  The tool works by taking the user's input query, and reasonable alternative classes that the user thinks could cause classification confusion. If the user does not know what alternatives to select, we provide an option to generate them using the following prompt to OpenAI's GPT-4o model to generate alternative labels: ``Create 10 realistic alternatives to the following input label by suggesting alternatives that are somewhat similar. I don't want the same label reworded or a subclass of the input label. The given label is: \{input\}''. The user can also input a domain so that the LLM can disambiguate the class more accurately. We generate images of the input query as well as the alternatives, compute the compound scores, and map the score into an accuracy prediction.

\subsection{Case Study}
There are many potential use cases for wanting to predict the zero-shot capabilities of a VLM on any arbitrary query.  Here we consider a specific case study that highlights how it might be used to differentiate between species of related looking insects. 

\begin{figure}[t]
\begin{center}
    \includegraphics[width=\columnwidth]{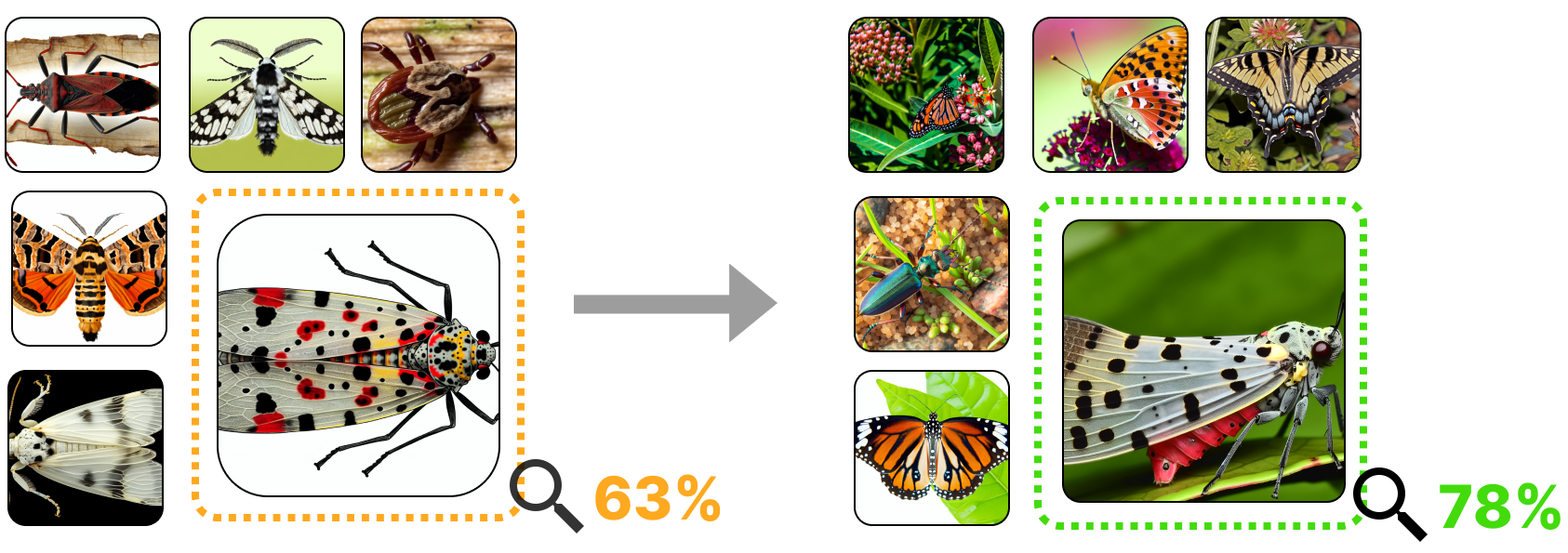}
\end{center}
\caption[Demo]{Our tool predicts that CLIP would be able to correctly classify images of spotted lanternflies relative to images of similar-looking insects about 63\% predicted accuracy. However, when the alternative classes are endangered and other threatened species in the northeast United States, CLIP is predicted to correctly classify a spotted lanternfly relative to the other insects about with 78\% predicted accuracy. }
\label{fig:lanternflyFig}
\end{figure}

The spotted lanternfly is an invasive species in the northeast of the United States that threatens to destroy local vegetation~\cite{slf-arxiv}. The USDA recommends killing, photographing and reporting the insect~\cite{USDA_SLFL}.  But how easy is it to automatically recognize these insects? Our tool suggests that CLIP has reasonable accuracy in differentiating Spotted Lanternflies from similar looking insects.  Figure~\ref{fig:lanternflyFig} (left) shows the spotted lanternfly compared to five similar looking insects.

An alternative app, however, might seek to ensure that someone does not accidentally kill endangered insects.  Figure~\ref{fig:lanternflyFig} (right) shows our tool predicting that CLIP can differentiate Spotted Lanternflies from these endangered species even more accurately. This shows how different users might iterate over queries to get more or less specific depending on their use case. They could also specify the habitat of each insect for the photos making the classification problem even more specific. Most importantly, they are getting feedback from the model in the form of the predicted accuracy and the images which add context to how our method got to its predicted accuracy, potentially increasing user trust in the system.

\section{Conclusions}
In this paper, we presented an approach to understand how well a VLM model will perform on a users zero-shot classification task.  We believe that this approach can have practical benefits for non-experts deciding if a VLM is an appropriate tool for their problem domain.  Our work shows that substantial benefit can be gained from analyzing the embeddings of artificially generated imagery related to the task.

We also find that there are a few cases in which our scores are not good predictors of performance.  The most interesting of these are cases like ObjectNet, where the image data is not well described by a text description like "an image of a rocking chair". Future work should explore effective approaches to allow the user to describe more about the context and background and circumstances of their image, and then take that information into account when creating a prediction.

This material is based upon work supported in whole or in part with funding from the National Science Foundation (DGE-2125677, IIS-2441774) and the Department of Defense (DoD). Any opinions, findings, conclusions, or recommendations
expressed in this material are those of the author(s) and do not necessarily reflect the views of the DOD/NSF and/or any agency or entity of the United States Government.

\bibliographystyle{IEEEtran}
\bibliography{references        }

\clearpage

\appendix

\renewcommand{\thesection}{\Alph{section}}
\section{Appendix}

\subsection{Image Captioning}
\label{sec:imageCaptioning}
We generated image captions for each class in all datasets for use in image and text experiments. Our captioning approach involved taking the list of class names and the category of a dataset and using that information to prompt GPT-4o to come up with descriptive image captions that we can use to generate diverse realistic images. Our generation is a multi-stage process that gets descriptive content about about a class to generate a realistic, descriptive, and diverse caption then generate the image. Our prompting is as follows. 1st stage: "You are an AI assistant that generates creative and diverse image captions suitable for use with image generation models like DALL-E. Given a subject, provide {num\_captions} distinct, diverse and descriptive captions, considering the following global taxonomical traits when generating captions: \{global\_traits\}." 2nd stage: "Please generate {num\_captions} diverse and creative alternative captions for the subject '{subject}'. Each caption should be compatible with the CLIP model so your caption should share the same prefix with the original prompt template provided: '{prompt\_template}'. An example can be, the template is 'a photo of a {c}' and the descriptive caption is 'a photo of a {c}, [descriptive content]'. Here {c} is the class name of the subject."

\begin{figure*}
    \centering
        \includegraphics[width=0.95\textwidth]{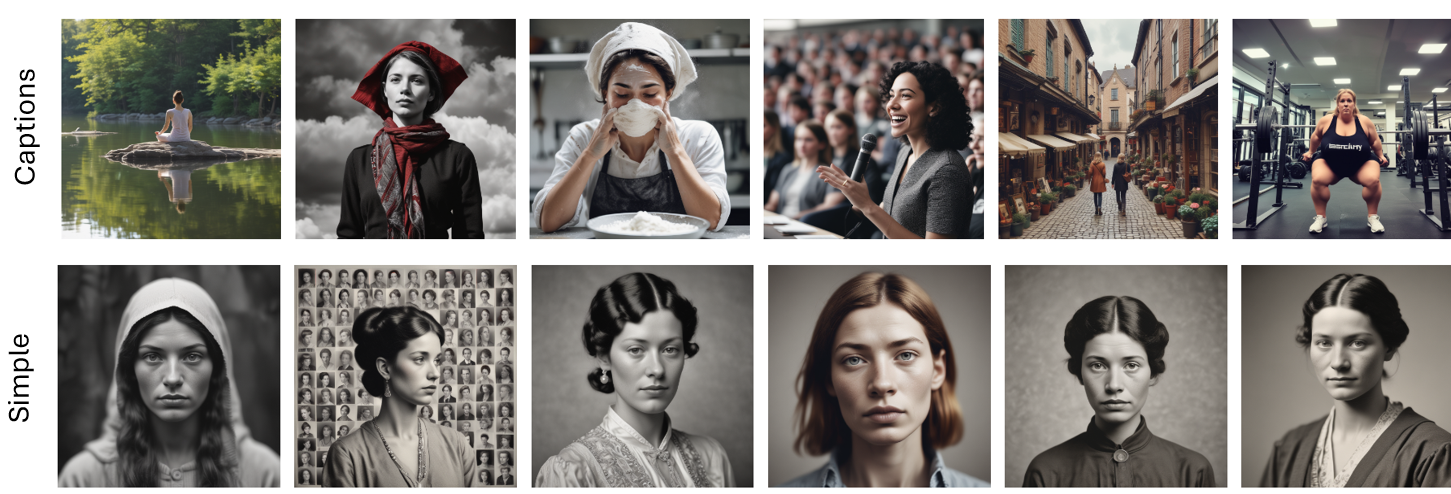}
        \caption[Generated Images]{A comparison of images generated from diverse text captions (top) and images generated with the simple prompt of the class name "woman" (bottom). The text captions lead to more detailed and diverse images of the class which matters much more in general classes such as "woman".}
    \label{fig:generatedImagesFig}
\end{figure*}

We find that images generated with the same non-descriptive class name prompt are considerably less diverse and look less realistic or only emulate one realistic setting. An example of this can be seen in Figure~\ref{fig:generatedImagesFig} with the class "Woman". The generative caption images show many diverse scenarios and characters that are all women, whereas the simply generated images ("a photo of a woman") almost look like the same image multiple times.

\begin{table}[]
    \centering
    \renewcommand{\arraystretch}{1.2}
    \resizebox{\columnwidth}{!}{%
        \begin{tabular}{|l|p{5cm}|}
            \hline
            \textbf{Class} & \textbf{Captions} \\ \hline
            Woman & A photo of the small woman exploring antique shops on a cobblestone street. \\ \hline
            Woman & A photo of the big woman effortlessly lifting weights in a gym. \\ \hline
            Woman & A photo of a woman wearing a traditional dress at a cultural festival. \\ \hline
            Woman & A photo of a woman hiking on a scenic mountain trail with her friend. \\ \hline
            Woman & A mysterious photo of a woman in a vintage dress standing in an old mansion. \\ \hline
            Black footed Albatross & A photo of a black footed Albatross, soaring gracefully above the ocean waves with its long wings spread wide. \\ \hline
            Black footed Albatross & A photo of a black footed Albatross, displaying its striking black and white plumage against the backdrop of a clear blue sky. \\ \hline
            Black footed Albatross & A photo of a black footed Albatross, nestled on a rocky cliff during the breeding season, showcasing its fluffy chick. \\ \hline
            Black footed Albatross & A photo of a black footed Albatross, expertly gliding over the surface of the water, its keen eyes scanning for fish. \\ \hline
            Black footed Albatross & A photo of a black footed Albatross, with its uniquely shaped beak, ready to dive for its next meal in the vast sea. \\ \hline
            Lasagna & A photo of a lasagna, a classic Italian dish layered with rich tomato sauce, creamy béchamel, and melted mozzarella. \\ \hline
            Lasagna & A photo of a lasagna, a comfort food favorite served hot, oozing with gooey cheese and savory meat. \\ \hline
            Lasagna & A photo of a lasagna, a delicious baked pasta dish with layers of succulent ground beef and tangy marinara sauce. \\ \hline
            Lasagna & A photo of a lasagna, an indulgent layered casserole with a golden-brown crust and fragrant herbs throughout. \\ \hline
            Lasagna & A photo of a lasagna, a rustic Italian family meal traditionally prepared in a large, deep dish. \\ \hline
            F-16A/B & A photo of the F-16A/B, a type of aircraft, displayed on a military base with combat-ready preparations. \\ \hline
            F-16A/B & A photo of a F-16A/B, a type of aircraft, in a dogfight simulation with other fighter jets. \\ \hline
            F-16A/B & A photo of the F-16A/B, a type of aircraft, during sunset, highlighting its silhouette against the vibrant colors of the sky. \\ \hline
            F-16A/B & A photo of a F-16A/B, a type of aircraft, equipped with precision-guided munitions, ready for a mission. \\ \hline
            F-16A/B & A photo of the F-16A/B, a type of aircraft, soaring above mountainous terrain in a training exercise. \\ \hline
        \end{tabular}
    }
    \caption{Caption descriptions for different categories.}
    \label{tab:caption_table}
\end{table}

\subsection{Converting Scores to Accuracy Predictions}
\label{sec:scoreConversion}
Our Experimental Results show that the scores above correlate with zero-shot classification accuracy but do not directly predict it. On our interactive website, however, we want to give end users a prediction of accuracy rather than correlations. In order to do this, we use beta regression to transform predicted scores into accuracy values.

Given $N$ datasets, let $\mathbf{S}_i$ and $\mathbf{A}_i$ represent the scores and true accuracies for dataset $i$. $\mathbf{S}_i$ may include any relevant scores. We evaluate using leave-one-out cross-validation, training on $N-1$ datasets and testing on the left-out dataset.

In the beta regression model, accuracy $A_j \in (0, 1)$ for class $j$ follows a beta distribution:
\[
A_j \sim \text{Beta}(\alpha_j, \beta_j),
\]
where $\alpha_j$ and $\beta_j$ are functions of the score $S_j$:
\[
\log(\alpha_j) = \mathbf{X}_j \cdot \boldsymbol{\theta}_1, \quad \log(\beta_j) = \mathbf{X}_j \cdot \boldsymbol{\theta}_2.
\]
Here, $\mathbf{X}_j$ is a feature vector derived from $S_j$, and $\boldsymbol{\theta}_1, \boldsymbol{\theta}_2$ are regression parameters.


    

\begin{table*}
    \centering
    \caption{Spearman Correlation for Text-based Scores}
    \label{tab:text_results}
    \resizebox{\textwidth}{!}{%
    \begin{tabular}{l|ccccccccc|}
    \cline{2-10}
    & \multicolumn{3}{c|}{Consistency} & \multicolumn{3}{c|}{Silhouette} & \multicolumn{3}{c|}{Compound} \\ \hline
    \multicolumn{1}{|l|}{Dataset} & CLIP & SigLIP & \multicolumn{1}{c|}{FLAVA} & CLIP & SigLIP & \multicolumn{1}{c|}{FLAVA} & CLIP & SigLIP & FLAVA \\ \hline
    \multicolumn{1}{|l|}{CUB}       & \colorcorr{0.50}  & \colorcorr{0.31}  & \multicolumn{1}{c|}{\colorcorr{0.57}}   & \colorcorr{0.50}  & \colorcorr{0.31}  & \multicolumn{1}{c|}{\colorcorr{0.51}}   & \colorcorr{0.52}  & \colorcorr{0.32}  & \colorcorr{0.56}  \\
    \multicolumn{1}{|l|}{Cars}      & \colorcorr{0.51}  & \colorcorr{0.35}  & \multicolumn{1}{c|}{\colorcorr{0.55}}   & \colorcorr{0.52}  & \colorcorr{0.34}  & \multicolumn{1}{c|}{\colorcorr{0.59}}   & \colorcorr{0.52}  & \colorcorr{0.36}  & \colorcorr{0.57}  \\
    \multicolumn{1}{|l|}{Aircraft}  & \colorcorr{0.35}  & \colorcorr{0.62}  & \multicolumn{1}{c|}{\colorcorr{0.21}}   & \colorcorr{0.35}  & \colorcorr{0.55}  & \multicolumn{1}{c|}{\colorcorr{0.21}}   & \colorcorr{0.36}  & \colorcorr{0.60}  & \colorcorr{0.21}  \\
    \multicolumn{1}{|l|}{Food}      & \colorcorr{0.31}  & \colorcorr{0.26}  & \multicolumn{1}{c|}{\colorcorr{0.20}}   & \colorcorr{0.35}  & \colorcorr{0.21}  & \multicolumn{1}{c|}{\colorcorr{-0.09}}  & \colorcorr{0.34}  & \colorcorr{0.22}  & \colorcorr{-0.02} \\
    \multicolumn{1}{|l|}{ImageNet}  & \colorcorr{0.27}  & \colorcorr{0.19}  & \multicolumn{1}{c|}{\colorcorr{0.24}}   & \colorcorr{0.30}  & \colorcorr{0.14}  & \multicolumn{1}{c|}{\colorcorr{0.20}}   & \colorcorr{0.29}  & \colorcorr{0.17}  & \colorcorr{0.22}  \\
    \multicolumn{1}{|l|}{ObjectNet} & \colorcorr{-0.07} & \colorcorr{-0.02} & \multicolumn{1}{c|}{\colorcorr{0.06}}   & \colorcorr{0.08}  & \colorcorr{-0.01} & \multicolumn{1}{c|}{\colorcorr{-0.07}}  & \colorcorr{0.01}  & \colorcorr{-0.02} & \colorcorr{-0.02} \\
    \multicolumn{1}{|l|}{CIFAR}     & \colorcorr{-0.15} & \colorcorr{-0.09} & \multicolumn{1}{c|}{\colorcorr{-0.14}}  & \colorcorr{-0.06} & \colorcorr{-0.02} & \multicolumn{1}{c|}{\colorcorr{0.24}}   & \colorcorr{-0.10} & \colorcorr{-0.03} & \colorcorr{0.12}  \\
    \multicolumn{1}{|l|}{Flowers}   & \colorcorr{0.55}  & \colorcorr{0.11}  & \multicolumn{1}{c|}{\colorcorr{0.42}}   & \colorcorr{0.56}  & \colorcorr{0.08}  & \multicolumn{1}{c|}{\colorcorr{0.26}}   & \colorcorr{0.60}  & \colorcorr{0.10}  & \colorcorr{0.34}  \\
    \multicolumn{1}{|l|}{Pets}      & \colorcorr{0.27}  & \colorcorr{0.21}  & \multicolumn{1}{c|}{\colorcorr{0.20}}   & \colorcorr{0.39}  & \colorcorr{-0.01} & \multicolumn{1}{c|}{\colorcorr{-0.25}}  & \colorcorr{0.37}  & \colorcorr{0.06}  & \colorcorr{-0.18} \\
    \multicolumn{1}{|l|}{RESISC}    & \colorcorr{0.31}  & \colorcorr{0.28}  & \multicolumn{1}{c|}{\colorcorr{0.14}}   & \colorcorr{0.34}  & \colorcorr{0.13}  & \multicolumn{1}{c|}{\colorcorr{0.29}}   & \colorcorr{0.34}  & \colorcorr{0.21}  & \colorcorr{0.24}  \\ \hline
    \end{tabular}
    }
\end{table*}

\subsection{Spearman Correlation for Text-based Scores}
Table~\ref{tab:text_results} displays the spearman correlations between the text consistency, silhouette, and compound scores to the accuracy of the real images. These score demonstrate how captions are generally weaker image proxies than synthetic images. The scores tend to follow similar patterns to the image results, but with most correlations being far lower. The CIFAR dataset is notably far worse than its corresponding image scores perhaps indicating how the captions fail to capture the more general classes of the dataset compared to the lower quality real images.

\subsection{Reproducibility}
 Datasets can be accessed at: https://huggingface.co/clip-benchmark. The evaluation of real and generated data for a single dataset took between 10 and 25 minutes running on an Nvidia GeForce RTX 3080 depending on the size of the dataset. The code to reproduce this result is at: https://github.com/Will-It-Zero-Shot/Will-it-Zero-Shot.
\end{document}